\documentclass[a4paper]{article}

\usepackage{INTERSPEECH2022}
\usepackage{bbold}
\usepackage{multirow}
\usepackage{makecell}
\usepackage{cite}
\usepackage{caption}
\usepackage{subcaption}
\usepackage{amsmath}
\usepackage{enumitem}

\title{Improved Consistency Training for Semi-Supervised Sequence-to-Sequence ASR via Speech Chain Reconstruction and Self-Transcribing}
\name{Heli Qi$^1$, Sashi Novitasari$^1$, Sakriani Sakti$^2$, Satoshi Nakamura$^1$}
\address{
  $^1$Nara Institute of Science and Technology, Japan\\
  $^2$Japan Advanced Institute of Science and Technology, Japan}
\email{\{qi.heli.qi9, sashi.novitasari.si3, s-nakamura\}@is.naist.jp, ssakti@jaist.ac.jp}

\begin{document}

\maketitle
\begin{abstract}
Consistency regularization has recently been applied to semi-supervised sequence-to-sequence (S2S) automatic speech recognition (ASR). This principle encourages an ASR model to output similar predictions for the same input speech with different perturbations. The existing paradigm of semi-supervised S2S ASR utilizes SpecAugment as data augmentation and requires a static teacher model to produce pseudo transcripts for untranscribed speech. However, this paradigm fails to take full advantage of consistency regularization. First, the masking operations of SpecAugment may damage the linguistic contents of the speech, thus influencing the quality of pseudo labels. Second, S2S ASR requires both input speech and prefix tokens to make the next prediction. The static prefix tokens made by the offline teacher model cannot match dynamic pseudo labels during consistency training. In this work, we propose an improved consistency training paradigm of semi-supervised S2S ASR. We utilize speech chain reconstruction as the weak augmentation to generate high-quality pseudo labels. Moreover, we demonstrate that dynamic pseudo transcripts produced by the student ASR model benefit the consistency training. Experiments on LJSpeech and LibriSpeech corpora show that compared to supervised baselines, our improved paradigm achieves a 12.2\% CER improvement in the single-speaker setting and 38.6\% in the multi-speaker setting. 
\end{abstract}

\section{Introduction}
In recent years, sequence-to-sequence (S2S) ASR has made significant progress thanks to the advancement of deep neural networks. S2S ASR models are designed for directly converting the input speech into transcripts \cite{chan2016listen,bahdanau2016end,chorowski2015attention}. However, a large amount of transcribed speech data is essential for training S2S ASR models to achieve state-of-the-art performance. Thus, many semi-supervised learning algorithms have been proposed to efficiently train ASR models with the help of untranscribed speech \cite{zhang2020semi,weninger2020semi,masumura2020sequence,chen2021semi,kahn2020self,park2020improved,higuchi2021momentum,xiao2021contrastive}.

Consistency regularization \cite{bachman2014learning} is an important principle of semi-supervised learning algorithms. This principle was originally designed for semi-supervised image classification \cite{samuli2017temporal,sajjadi2016regularization,sohn2020fixmatch,berthelot2019mixmatch} and it has recently been extended to semi-supervised S2S ASR \cite{zhang2020semi,weninger2020semi,masumura2020sequence,chen2021semi,wang2020improving}. Consistency regularization assumes that an ASR model should output similar predictions for the same input speech with various perturbations. Also, these perturbations should change the distribution of input speech without altering the corresponding transcripts\cite{sohn2020fixmatch}.
In the literature, SpecAugment \cite{park2019specaugment} is commonly used to perturb speech features due to its simplicity. Its time-frequency masking plays a major role in improving the robustness of ASR \cite{wang2020improving}. However, randomly removing continuous frequency bins or temporal frames may damage the semantics of the input speech. Incomplete linguistic contents will further accumulate errors in pseudo labels and thus influence ASR performance during consistency training. 

Different from image classification, S2S ASR models require both input speech and prefix tokens to make the prediction at each time step. In the consistency training paradigm for S2S ASR \cite{weninger2020semi,chen2021semi,zhang2020semi,masumura2020sequence,park2020improved,kahn2020self}, a teacher model trained on transcribed speech is used to produce pseudo transcripts for untranscribed speech. The pseudo transcripts are then fed into the student ASR model as prefix tokens during consistency training. However, there exist some errors in predictions of the teacher model because of the limited training set \cite{chen2021semi}. These errors remain in the static pseudo transcripts and make pseudo labels poorer in quality. Moreover, since the static pseudo transcripts are made by original speech before consistency training, the mismatch between these transcripts and perturbed speech will further influence the student model during consistency training.

This paper presents an improved consistency training paradigm of S2S ASR. In previous work, machine speech chain \cite{tjandra2017listening,tjandra2020machine} was designed to jointly train ASR and TTS by reconstructing unlabeled speech and text data. 
In this work, we adopt the speech chain reconstruction as a data augmentation method and focus on the FixMatch algorithm \cite{sohn2020fixmatch} which has recently been applied on S2S ASR \cite{weninger2020semi}. Our contributions are as follow:
\begin{itemize}
\item We propose the self-transcribing scheme where the student model serves as its own teacher. Specifically, the pseudo transcripts are produced dynamically by the student model with perturbed speech as input.
\item We demonstrate that speech chain reconstruction is superior than SpecAugment as the weak augmentation for making pseudo labels.
\item We conducted constrast experiments that covers most of the factors that may have an impact on the consistency training, including number of speakers, confidence threshold, ratio of unlabeled data to labeled data, data augmentation methods, and pseudo transcript generation.
\end{itemize}

\section{Semi-supervised consistency training for S2S ASR}
\begin{figure*}
  \centering
  \begin{subfigure}[b]{0.3\textwidth}
    \centering
    \includegraphics[width=\textwidth]{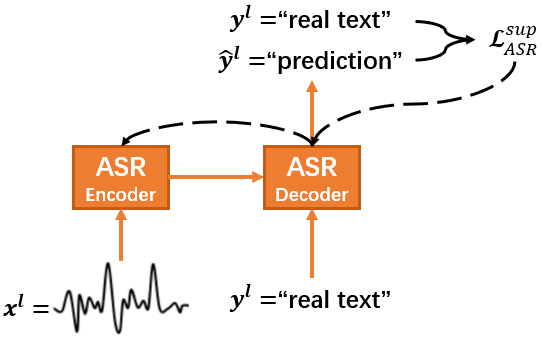}
    \caption{\label{fig:supervised}}
  \end{subfigure}
  \begin{subfigure}[b]{0.6\textwidth}
    \centering
    \includegraphics[width=\textwidth]{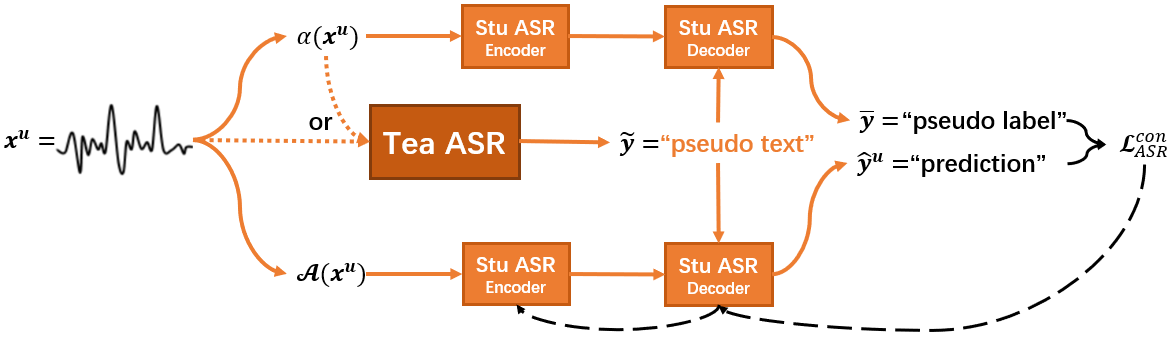}
    \caption{\label{fig:consistency}}
  \end{subfigure}
  \label{fig:paradigm}
  \caption{``Tea'' denotes the teacher model and ``Stu'' denotes the student model. Black dash lines represent the back-propagation of the gradients. (a) Supervised training for sequence-to-sequence ASR. (b) FixMatch-based consistency training for sequence-to-sequence ASR. Orange dash lines represent the different input speech for making pseudo transcripts.}
\end{figure*}

\subsection{Supervised training for a base ASR model}
S2S ASR models are designed to directly predict the conditional probability $P(\bm{\hat{y}}|\bm{x}; \theta_{ASR})$ of a sequence of predicted tokens $\bm{\hat{y}} = [\hat{y}_1,...,\hat{y}_T]$ given a sequence of speech features $\bm{x} = [x_1,...,x_S]$. Here, $\theta_{ASR}$ represents the parameters of ASR, $S$ is the length of the input sequence, and $T$ is the length of the output sequence. Our ASR models are based on Listen-Attend-Spell (LAS) \cite{chan2016listen} which has an encoder-decoder architecture as shown in Fig.\ref{fig:supervised}.

The encoder converts the input sequence of speech features into a sequence of hidden representations $\bm{h}^e = [h^e_1,...,h^e_S]$. The decoder receives $\bm{h}^e$ and outputs the token probability $p(\hat{y}_t|\hat{y}_{1:t-1}, \bm{x}; \theta_{ASR})$ at time step $t$ based on the prefix tokens $\hat{y}_{1:t-1}$. The probability of the generated transcript $P(\bm{\hat{y}}|\bm{x}; \theta_{ASR})$ is calculated by the product of the token probability at each time step as
\begin{equation}
  P(\bm{\hat{y}}|\bm{x}; \theta_{ASR}) = \prod_{t=1}^T{p(\hat{y}_t|\hat{y}_{1:t-1}, \bm{x}; \theta_{ASR})}.
  \label{eq1}
\end{equation}

During training, the prefix tokens $\hat{y}_{1:t-1}$ are replaced with the ground-truth labels. Given a speech-text pair $(\bm{x^l}, \bm{y^l})$, a base ASR model is trained by the following supervised ASR loss:
\begin{equation}
  \mathcal{L}^{sup}_{ASR} = - \frac{1}{T} \sum_{t=1}^T{\log{p(y^l_t|y^l_{1:t-1}, \bm{x}; \theta_{ASR})}}.
  \label{eq2}
\end{equation}




\subsection{Consistency training for S2S ASR based on FixMatch algorithm \cite{weninger2020semi}}
FixMatch algorithm \cite{sohn2020fixmatch} is a semi-supervised algorithm for image classification that combines consistency regularization with pseudo-labeling. In its existing application to S2S ASR \cite{weninger2020semi}, a teacher ASR model is initialized on the parameters of the base ASR model and used to produce the pseudo transcript $\bm{\Tilde{y}}$ for the untranscribed speech $\bm{x^u}$ as
\begin{equation}
  \bm{\Tilde{y}} = \mathop{\arg\max}\limits_{\bm{z}} P(\bm{z}|\bm{x^u}; \theta^{Tea}_{ASR}).
  \label{eq3}
\end{equation}

As shown in Fig.\ref{fig:consistency}, untranscribed speech $\bm{x^u}$ is separately perturbed by a weak augmentation function $\alpha(\cdot)$ and a strong augmentation function $\mathcal{A}(\cdot)$. The weakly-perturbed speech $\alpha(\bm{x^u})$ is fed into the student ASR model to obtain the pseudo label $\bar{y}_t$ at time step $t$ given $\Tilde{y}_{1:t-1}$ as prefix tokens by
\begin{gather}
    \bar{y}_t = \mathop{\arg\max}\limits_{z} p(z|\Tilde{y}_{1:t-1}, \alpha(\bm{x^u}); \theta^{Stu}_{ASR}).
\end{gather}

The consistency ASR loss $\mathcal{L}^{con}_{ASR}$ is calculated on the strongly-perturbed speech $\mathcal{A}(\bm{x^u})$ and pseudo label $\bar{y}_t$ given the same pseudo transcripts $\Tilde{y}_{1:t-1}$ as prefix tokens by 
\begin{equation}
  \mathcal{L}^{con}_{ASR} = - \frac{1}{T} \sum_{t=1}^T \mathbb{1}(q_t > \tau) \log{p(\bar{y}_t|\Tilde{y}_{1:t-1}, \mathcal{A}(\bm{x^u}); \theta^{Stu}_{ASR})},
\label{eq5}
\end{equation}
where $q_t = p(\bar{y}_t|\Tilde{y}_{1:t-1}, \alpha(\bm{x^u}); \theta^{Stu}_{ASR})$ is the confidence of the student model on $\bar{y}_t$ and $\tau$ is the confidence threshold. 

Finally, the student model is trained by the following loss function: 
\begin{equation}
  \mathcal{L}_{ASR} = \mathcal{L}^{sup}_{ASR} + \lambda_{con}\mathcal{L}^{con}_{ASR}.
\label{eq6}
\end{equation}


\subsection{Proposed self-transcribing scheme}
On top of the existing paradigm, we made two improvements. First, the weakly-perturbed speech $\alpha(\bm{x^u})$ is used to produce the pseudo transcripts $\bm{\Tilde{y}}$. Second, the student model is initialized by the base ASR and dynamically produces the pseudo transcripts by itself during the consistency training. Our improvements can be formualted as follow:
\begin{equation}
  \bm{\Tilde{y}} = \mathop{\arg\max}\limits_{\bm{z}} P(\bm{z}|\alpha(\bm{x^u}); \theta^{Stu}_{ASR}).
\label{eq7}
\end{equation}


\section{Speech chain reconstruction}
\subsection{Semi-supervised TTS based on pseudo transcribing}
Sequence-to-sequence TTS models can be considered the reverse case of ASR that directly predicts the conditional probability $P(\bm{\hat{x}}|\bm{y}, \theta_{TTS})$ of a sequence of speech features $\bm{\hat{x}} = [\hat{x}_1, ..., \hat{x}_S]$ given a sequence of tokens $\bm{y} = [y_1, ..., y_T]$. Our TTS model is based on Tacotron2 \cite{shen2018natural} which has a similar architecture to LAS. We provide the decoder of TTS with the speaker embedding $f(\bm{x})$ extracted from the input speech, which enables TTS to synthesize speech in a multi-speaker setting. Our TTS model is trained by the loss function $\mathcal{L}_{TTS}$:
\begin{flalign}
  && \mathcal{L}_{TTS} = & \mathcal{L}^{sup}_{TTS} + \mathcal{L}^{unsup}_{TTS}, &\\
  && \mathcal{L}^{sup}_{TTS} = & \frac{1}{S} \sum_{s=1}^S (x^l_s - \hat{x}^l_s)^2 & \nonumber\\
  && & - (b^l_s\log(\hat{b}^l_s) + (1 - b^l_s)\log(1 - \hat{b}^l_s)),  &\\
  && \mathcal{L}^{unsup}_{TTS} = & \frac{1}{S} \sum_{s=1}^S (x^u_s - \hat{x}^u_s)^2 & \nonumber\\
  && & - (b^u_s\log(\hat{b}^u_s) + (1 - b^u_s)\log(1 - \hat{b}^u_s)), &
\end{flalign}
where $\hat{x}_s, x_s$ are the predicted and ground-truth log Mel-scale spectrograms at time $s$ and $\hat{b}_s, b_s$ are the predicted and ground-truth end-of-frame probabilities. The pseudo transcripts $\bm{\Tilde{y}}$ used for training the TTS model is produced by the base ASR model before ASR consistency training as Eq.\ref{eq3} does.

\subsection{Speech chain reconstruction for untranscribed speech}
\begin{figure}[t]
  \centering
  \includegraphics[width=0.8\linewidth]{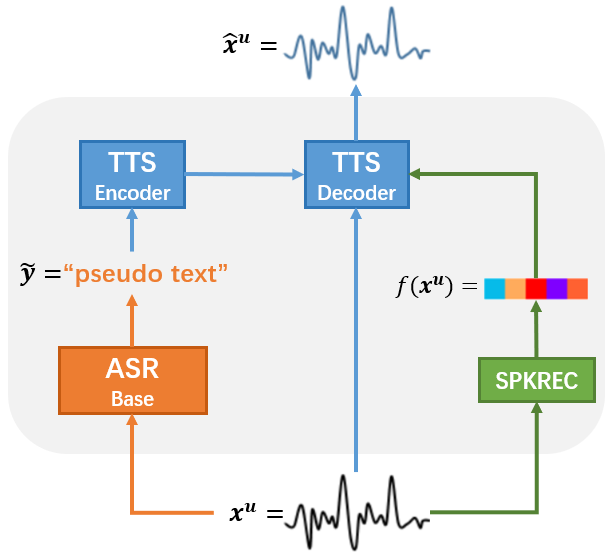}
  \caption{Proposed speech chain reconstruction. The grey box can be viewed as the weak augmentation $\alpha(\cdot)$.}
  \label{fig:speechain}
\end{figure}
Different from our previous work that generates synthetic speech from real texts for semi-supervised ASR training \cite{tjandra2017listening,tjandra2020machine}, this work applied the speech chain reconstruction as a data augmentation method for the untranscribed speech by the teacher-forcing technique. This technique prevents TTS decoding from mispronunciation and early stopping, thus protecting the linguistic contents. Fig.\ref{fig:speechain} shows the process of our proposed speech chain reconstruction that can be considered as a frame-level speech transformation. 

Untranscribed speech $\bm{x^u}$ is utilized three times before generating reconstructed speech $\bm{\hat{x}^u}$. First, the base ASR model converts $\bm{x^u}$ into pseudo transcripts $\bm{\Tilde{y}}$ as the input of our TTS encoder. Then, the speaker embedding $f(\bm{x^u})$ is extracted from $\bm{x^u}$ to provide our TTS decoder with the speaker information. Finally, our TTS decoder generates the reconstructed speech $\bm{\hat{x}^u}$ with original speech $\bm{x^u}$ as the prefix speech features at each time step. The reconstructed speech $\bm{\hat{x}^u}$ is treated as the weakly-perturbed speech $\alpha(\bm{x^u})$ during the subsequent ASR consistency training.

\section{Experiments}

\subsection{Datasets}
We conducted experiments in both single-speaker and multi-speaker settings. We use the LJSpeech corpus \cite{ljspeech17} for the single-speaker setting. We take 12,600 utterances as the training set, 250 utterances as the development set and 250 utterances as the test set. There is no overlap in the division of data. We treat the first 50\% of the training set as labeled data and the last 50\% as unlabeled data. In the multi-speaker setting, we use the LibriSpeech corpus \cite{panayotov2015librispeech}. We take the ``train-clean-100" set as the training set, the ``dev-clean" set as the development set and the ``test-clean" set as the test set. We treat the first 75 speakers (8,570 utts) of the training set as labeled data and the last 176 speakers (19,968 utts) as unlabeled data.

As for acoustic features, we used 80-dimensional log Mel-spectrograms extracted at a 50-ms frame length and a 12.5-ms frame shift. The English letters in the transcripts of all utterances are normalized into their lowercase forms. All transcripts are mapped into a 29-token set: 26 (a-z) letters of the alphabet, apostrophes, space, and ``\emph{sos}/\emph{eos}" $\footnote{We combined \emph{sos} and \emph{eos} into one token.}$. 

\subsection{Model configuration}
\subsubsection{ASR}
In the single-speaker setting, our ASR encoder is composed of three bidirectional LSTM layers with 256 hidden units for each direction (totally 512 hidden units for each Bi-LSTM layer). Hierarchical sub-sampling \cite{chan2016listen,bahdanau2016end} was used on the last two layers to reduce the sequence length by a factor of four. In the multi-speaker setting, we added two extra Bi-LSTM layers at the beginning of the encoder. On the decoder sides, the single-speaker and multi-speaker settings shared the same configuration: an embedding layer followed by a unidirectional LSTM layer with 512 hidden units. We selected Additive Attention \cite{bahdanau2015neural} as the attention module of our ASR models. The beam searching technique was adopted to generate pseudo transcripts during the consistency training and the beam size was set to 4.

AdaDelta \cite{zeiler2012adadelta} was adopted to train our ASR models. During the base training, the initial learning rate was set to 1.0 and the decay rate was set to 0.1. During the consistency training, we set a smaller initial learning rate of 0.5 and a larger decay rate of 0.2. The learning rate decay was based on the accuracy calculated on the development set during training and the minimal learning rate was set to 1\% of the initial value. We used early stopping to prevent the models from overfitting. The weight of the consistency loss $\mathcal{L}^{con}_{ASR}$ is set to 0.1.


\subsubsection{TTS}
The hyperparameters for our TTS model were generally the same as those for the original Tacotron2, except we concatenated the encoder hidden representations with the speaker embedding vectors in the multi-speaker experiments. We extracted X-vectors \cite{snyder2018x} from the input speech as the speaker embedding vectors and the extraction was done using SpeechBrain \cite{speechbrain}.

Adam \cite{kingma2014adam} was adopted to train our TTS model with an initial learning rate of 0.001 and a decay rate of 0.1. The learning rate decay was based on the loss calculated on the development set during training and the minimal learning rate was set to 1\% of the initial value. Early stopping was also used to avoid overfitting. 
For each time-step, our model generated two consecutive frames to reduce the number of steps in the decoding process.

\subsection{Experimental setting}
ESPNET2 \cite{watanabe2018espnet,hayashi2020espnet} was used to perform our experiments. Our supervised baselines were the base ASR models trained using only the labeled data. We designed four scenarios and conducted contrast experiments where the weak augmentation $\alpha(\cdot)$ was either SpecAugment or speech chain reconstruction. For all experiments, the strong augmentation $\mathcal{A}(\cdot)$ was implemented by SpecAugment. Our weak SpecAugment applied one time of time-frequency masking while the strong SpecAugment applied two times. Maximal frequency masking width was 5 bins for $\alpha(\cdot)$ and 20 bins for $\mathcal{A}(\cdot)$. For the LJSpeech corpus, maximal time masking width was 10 frames for $\alpha(\cdot)$ and 50 frames for $\mathcal{A}(\cdot)$. For LibriSpeech corpus, maximal time masking width was 20 frames for $\alpha(\cdot)$ and 100 frames for $\mathcal{A}(\cdot)$.

\begin{table*}[t]
  \caption{CER results of our constrast experiments. Underlined numbers denote the best performance in each scenario.}
  \label{tab:single-speaker exp}
  \centering
  \begin{tabular}{c|ccccc|ccc}
    \toprule
     & \multicolumn{5}{c}{LJSpeech} & \multicolumn{3}{c}{LibriSpeech}\\
    $\alpha(\cdot)$ & $\tau$=0.5 & $\tau$=0.6 & $\tau$=0.7 & $\tau$=0.8 & $\tau$=0.9 & $\tau$=0.5 & $\tau$=0.7 & $\tau$=0.9\\
    \midrule
    \midrule
    \multicolumn{9}{c}{\emph{Supervised Baseline}}\\
    \textbf{--} & 8.2 & 8.2 & 8.2 & 8.2 & 8.2 & 28.0 & 28.0 & 28.0\\
    \midrule
    \midrule
    \multicolumn{9}{c}{\emph{Static $\bm{\Tilde{y}}$ produced by $\bm{x^u}$ (the existing paradigm~\cite{weninger2020semi})}}\\
    Weak SpecAugment & 8.3 & 7.8 & 7.7 & 7.5 & 7.7 & 18.3 & 19.6 & 20.8\\
    Speech Chain Reconstruction & 7.9 & 7.6 & \underline{7.4} & 7.5 & 7.8 & \underline{18.2} & 19.8 & 18.5\\
    \midrule
    \midrule
    \multicolumn{9}{c}{\emph{Static $\bm{\Tilde{y}}$ produced by $\alpha(\bm{x^u})$}}\\
    Weak SpecAugment & 7.8 & 7.7 & 7.7 & 7.8 & 7.6 & 18.8 & 19.6 & 20.3\\
    Speech Chain Reconstruction & 7.9 & 7.7 & \underline{7.2} & \underline{7.2} & 7.6 & \underline{17.2} & 18.4 & 18.3\\
    \midrule
    \midrule
    \multicolumn{9}{c}{\emph{Dynamic $\bm{\Tilde{y}}$ produced by $\bm{x^u}$}}\\
    Weak SpecAugment & 7.9 & 7.9 & 7.6 & 7.4 & 7.6 & 19.1 & 19.1 & 19.9\\
    Speech Chain Reconstruction & 8.2 & \underline{7.2} & 7.5 & 7.4 & 7.6 & \underline{18.1} & 18.4 & 18.4\\
    \midrule
    \midrule
    \multicolumn{9}{c}{\emph{Dynamic $\bm{\Tilde{y}}$ produced by $\alpha(\bm{x^u})$}}\\
    Weak SpecAugment & 7.5 & 7.4 & 8.1 & 7.6 & 8.0 & 19.8 & 20.5 & 18.9\\
    Speech Chain Reconstruction & 7.7 & 7.7 & 7.6 & 7.4 & \underline{7.2} & 20.0 & 19.1 & \underline{18.5}\\
    \bottomrule
  \end{tabular}
\end{table*}

\section{Results and analysis}
\subsection{Single-speaker experiments}
In the single-speaker setting, the ratio of labeled data to unlabeled data was set to 1:1. From Tab.\ref{tab:single-speaker exp}, we observed that speech chain reconstruction outperforms the weak SpecAugment in all scenarios. Our consistency training paradigm achieved the best CER performance of 7.2\%, which has a 12.2\% CER improvement on the supervised baseline. This showcases how speech chain reconstruction keeps more linguistic information than SpecAugment, and thus it is more suitable to be the weak augmentation.

In the single-speaker setting, dynamic pseudo transcripts produced by the student model achieved a 2.7\% CER improvement on the existing paradigm. This indicates that the student model gradually corrects the errors in the pretrained base model during consistency training. 
It can be seen that the pseudo transcripts produced by the weakly-perturbed input speech $\alpha(\bm{x^u})$ also resulted in a 2.7\% CER improvement. It supports our idea that prefix tokens should match the input speech during ASR consistency training. As for the reason why the relative improvements on the existing paradigm  were not large in scale, we hypothesized that the base model is already good enough to produce understandable transcripts because 50\% of the training set was used as the labeled data. 




\subsection{Multi-speaker experiments}

In the multi-speaker setting, we simulated a harsher condition where the ratio of labeled data to unlabeled data is 3:7. From Tab.\ref{tab:single-speaker exp}, speech chain reconstruction still outperformed the weak SpecAugment in all scenarios. With more unlabeled data, the improvement of our paradigm over the supervised baseline surged to 38.6\%, which indicates that our paradigm benefits from a large amount of untranscribed speech.

On top of the existing paradigm, a 5.5\% CER improvement was achieved when the input speech used to produce the pseudo transcripts was changed from the original speech $\bm{x^u}$ to weak-perturbed speech $\alpha(\bm{x^u})$. On the other hand, only a 0.5\% CER improvement was observed when we set the teacher model in the existing paradigm to the student model itself during consistency training. This indicates that the mismatch between the input speech and pseudo transcripts has a stronger influence on our ASR models than the errors in the pretrained base model.

According to the right part of Tab.\ref{tab:single-speaker exp}, our ASR models performed better when we set the confidence threshold to a smaller value. Since we only used the first 30\% of ``train-clean-100'' as the labeled data to train the base model, our student ASR models are not very confident on the untranscribed speech and thus output relatively lower token probability at each time step. With a higher confidence threshold, only a small fraction of untranscribed speech is utilized to calculate the final ASR loss, hence seriously restricting the potential of consistency training for improving ASR performance.

\section{Conclusions}
In this work, we proposed an improved consistency training paradigm for S2S ASR. We took the FixMatch algorithm as the proving ground and presented comprehensive constrast experiments covering most of the factors in the semi-supervised ASR training. Our results show that speech chain reconstruction protects more linguistic contents than SpecAugment and produces pseudo labels with higher quality. Moreover, the proposed self-transcribing method helps the student model correct the errors in the pretrained base model and eliminate the mismatch between the input speech and prefix tokens. Our future work involves applications of other semi-supervised algorithms on S2S ASR with various data augmentation methods for speech data.

\section{Acknowledgements}
Part of this work is supported by JSPS KAKENHI Grant Number JP21H05054 and JP21H03467.

~\\
~\\
~\\
~\\

\bibliographystyle{IEEEtran}

\bibliography{template}

\begin{thebibliography}{10}
\providecommand{\url}[1]{#1}
\csname url@samestyle\endcsname
\providecommand{\newblock}{\relax}
\providecommand{\bibinfo}[2]{#2}
\providecommand{\BIBentrySTDinterwordspacing}{\spaceskip=0pt\relax}
\providecommand{\BIBentryALTinterwordstretchfactor}{4}
\providecommand{\BIBentryALTinterwordspacing}{\spaceskip=\fontdimen2\font plus
\BIBentryALTinterwordstretchfactor\fontdimen3\font minus
  \fontdimen4\font\relax}
\providecommand{\BIBforeignlanguage}[2]{{%
\expandafter\ifx\csname l@#1\endcsname\relax
\typeout{** WARNING: IEEEtran.bst: No hyphenation pattern has been}%
\typeout{** loaded for the language `#1'. Using the pattern for}%
\typeout{** the default language instead.}%
\else
\language=\csname l@#1\endcsname
\fi
#2}}
\providecommand{\BIBdecl}{\relax}
\BIBdecl

\bibitem{chan2016listen}
W.~Chan, N.~Jaitly, Q.~Le, and O.~Vinyals, ``Listen, attend and spell: A neural
  network for large vocabulary conversational speech recognition,'' in
  \emph{2016 IEEE international conference on acoustics, speech and signal
  processing (ICASSP)}.\hskip 1em plus 0.5em minus 0.4em\relax IEEE, 2016, pp.
  4960--4964.

\bibitem{bahdanau2016end}
D.~Bahdanau, J.~Chorowski, D.~Serdyuk, P.~Brakel, and Y.~Bengio, ``End-to-end
  attention-based large vocabulary speech recognition,'' in \emph{2016 IEEE
  international conference on acoustics, speech and signal processing
  (ICASSP)}.\hskip 1em plus 0.5em minus 0.4em\relax IEEE, 2016, pp. 4945--4949.

\bibitem{chorowski2015attention}
J.~K. Chorowski, D.~Bahdanau, D.~Serdyuk, K.~Cho, and Y.~Bengio,
  ``Attention-based models for speech recognition,'' \emph{Advances in neural
  information processing systems}, vol.~28, 2015.

\bibitem{zhang2020semi}
Z.-q. Zhang, Y.~Song, J.-s. Zhang, I.~V. McLoughlin, and L.-r. Dai,
  ``Semi-supervised end-to-end asr via teacher-student learning with
  conditional posterior distribution.'' in \emph{INTERSPEECH}, 2020, pp.
  3580--3584.

\bibitem{weninger2020semi}
F.~Weninger, F.~Mana, R.~Gemello, J.~Andr{\'e}s-Ferrer, and P.~Zhan,
  ``Semi-supervised learning with data augmentation for end-to-end asr,'' in
  \emph{INTERSPEECH}, 2020.

\bibitem{masumura2020sequence}
R.~Masumura, M.~Ihori, A.~Takashima, T.~Moriya, A.~Ando, and Y.~Shinohara,
  ``Sequence-level consistency training for semi-supervised end-to-end
  automatic speech recognition,'' in \emph{ICASSP 2020-2020 IEEE International
  Conference on Acoustics, Speech and Signal Processing (ICASSP)}.\hskip 1em
  plus 0.5em minus 0.4em\relax IEEE, 2020, pp. 7054--7058.

\bibitem{chen2021semi}
Z.~Chen, A.~Rosenberg, Y.~Zhang, H.~Zen, M.~Ghodsi, Y.~Huang, J.~Emond,
  G.~Wang, B.~Ramabhadran, and P.~J.~M. Mengibar, ``Semi-supervision in asr:
  Sequential mixmatch and factorized tts-based augmentation,'' in
  \emph{INTERSPEECH}, 2021.

\bibitem{kahn2020self}
J.~Kahn, A.~Lee, and A.~Hannun, ``Self-training for end-to-end speech
  recognition,'' in \emph{ICASSP 2020-2020 IEEE International Conference on
  Acoustics, Speech and Signal Processing (ICASSP)}.\hskip 1em plus 0.5em minus
  0.4em\relax IEEE, 2020, pp. 7084--7088.

\bibitem{park2020improved}
D.~S. Park, Y.~Zhang, Y.~Jia, W.~Han, C.-C. Chiu, B.~Li, Y.~Wu, and Q.~V. Le,
  ``Improved noisy student training for automatic speech recognition,'' in
  \emph{INTERSPEECH}, 2020.

\bibitem{higuchi2021momentum}
Y.~Higuchi, N.~Moritz, J.~L. Roux, and T.~Hori, ``Momentum pseudo-labeling for
  semi-supervised speech recognition,'' in \emph{INTERSPEECH}, 2021.

\bibitem{xiao2021contrastive}
A.~Xiao, C.~Fuegen, and A.~Mohamed, ``Contrastive semi-supervised learning for
  asr,'' in \emph{ICASSP 2021-2021 IEEE International Conference on Acoustics,
  Speech and Signal Processing (ICASSP)}.\hskip 1em plus 0.5em minus
  0.4em\relax IEEE, 2021, pp. 3870--3874.

\bibitem{bachman2014learning}
P.~Bachman, O.~Alsharif, and D.~Precup, ``Learning with pseudo-ensembles,''
  \emph{Advances in neural information processing systems}, vol.~27, 2014.

\bibitem{samuli2017temporal}
L.~Samuli and A.~Timo, ``Temporal ensembling for semi-supervised learning,'' in
  \emph{International Conference on Learning Representations (ICLR)}, vol.~4,
  no.~5, 2017, p.~6.

\bibitem{sajjadi2016regularization}
M.~Sajjadi, M.~Javanmardi, and T.~Tasdizen, ``Regularization with stochastic
  transformations and perturbations for deep semi-supervised learning,''
  \emph{Advances in neural information processing systems}, vol.~29, 2016.

\bibitem{sohn2020fixmatch}
K.~Sohn, D.~Berthelot, N.~Carlini, Z.~Zhang, H.~Zhang, C.~A. Raffel, E.~D.
  Cubuk, A.~Kurakin, and C.-L. Li, ``Fixmatch: Simplifying semi-supervised
  learning with consistency and confidence,'' \emph{Advances in Neural
  Information Processing Systems}, vol.~33, pp. 596--608, 2020.

\bibitem{berthelot2019mixmatch}
D.~Berthelot, N.~Carlini, I.~Goodfellow, N.~Papernot, A.~Oliver, and C.~A.
  Raffel, ``Mixmatch: A holistic approach to semi-supervised learning,''
  \emph{Advances in Neural Information Processing Systems}, vol.~32, 2019.

\bibitem{wang2020improving}
G.~Wang, A.~Rosenberg, Z.~Chen, Y.~Zhang, B.~Ramabhadran, Y.~Wu, and P.~Moreno,
  ``Improving speech recognition using consistent predictions on synthesized
  speech,'' in \emph{ICASSP 2020-2020 IEEE International Conference on
  Acoustics, Speech and Signal Processing (ICASSP)}.\hskip 1em plus 0.5em minus
  0.4em\relax IEEE, 2020, pp. 7029--7033.

\bibitem{park2019specaugment}
D.~S. Park, W.~Chan, Y.~Zhang, C.-C. Chiu, B.~Zoph, E.~D. Cubuk, and Q.~V. Le,
  ``Specaugment: A simple data augmentation method for automatic speech
  recognition,'' \emph{Proc. Interspeech 2019}, pp. 2613--2617, 2019.

\bibitem{tjandra2017listening}
A.~Tjandra, S.~Sakti, and S.~Nakamura, ``Listening while speaking: Speech chain
  by deep learning,'' in \emph{2017 IEEE Automatic Speech Recognition and
  Understanding Workshop (ASRU)}.\hskip 1em plus 0.5em minus 0.4em\relax IEEE,
  2017, pp. 301--308.

\bibitem{tjandra2020machine}
------, ``Machine speech chain,'' \emph{IEEE/ACM Transactions on Audio, Speech,
  and Language Processing}, vol.~28, pp. 976--989, 2020.

\bibitem{shen2018natural}
J.~Shen, R.~Pang, R.~J. Weiss, M.~Schuster, N.~Jaitly, Z.~Yang, Z.~Chen,
  Y.~Zhang, Y.~Wang, R.~Skerrv-Ryan \emph{et~al.}, ``Natural tts synthesis by
  conditioning wavenet on mel spectrogram predictions,'' in \emph{2018 IEEE
  international conference on acoustics, speech and signal processing
  (ICASSP)}.\hskip 1em plus 0.5em minus 0.4em\relax IEEE, 2018, pp. 4779--4783.

\bibitem{ljspeech17}
K.~Ito and L.~Johnson, ``The lj speech dataset,''
  \url{https://keithito.com/LJ-Speech-Dataset/}, 2017.

\bibitem{panayotov2015librispeech}
V.~Panayotov, G.~Chen, D.~Povey, and S.~Khudanpur, ``Librispeech: an asr corpus
  based on public domain audio books,'' in \emph{2015 IEEE international
  conference on acoustics, speech and signal processing (ICASSP)}.\hskip 1em
  plus 0.5em minus 0.4em\relax IEEE, 2015, pp. 5206--5210.

\bibitem{bahdanau2015neural}
D.~Bahdanau, K.~H. Cho, and Y.~Bengio, ``Neural machine translation by jointly
  learning to align and translate,'' in \emph{3rd International Conference on
  Learning Representations, ICLR 2015}, 2015.

\bibitem{zeiler2012adadelta}
M.~D. Zeiler, ``Adadelta: an adaptive learning rate method,'' \emph{arXiv
  preprint arXiv:1212.5701}, 2012.

\bibitem{snyder2018x}
D.~Snyder, D.~Garcia-Romero, G.~Sell, D.~Povey, and S.~Khudanpur, ``X-vectors:
  Robust dnn embeddings for speaker recognition,'' in \emph{2018 IEEE
  international conference on acoustics, speech and signal processing
  (ICASSP)}.\hskip 1em plus 0.5em minus 0.4em\relax IEEE, 2018, pp. 5329--5333.

\bibitem{speechbrain}
M.~Ravanelli, T.~Parcollet, P.~Plantinga, A.~Rouhe, S.~Cornell, L.~Lugosch,
  C.~Subakan, N.~Dawalatabad, A.~Heba, J.~Zhong, J.-C. Chou, S.-L. Yeh, S.-W.
  Fu, C.-F. Liao, E.~Rastorgueva, F.~Grondin, W.~Aris, H.~Na, Y.~Gao, R.~D.
  Mori, and Y.~Bengio, ``{SpeechBrain}: A general-purpose speech toolkit,''
  2021, arXiv:2106.04624.

\bibitem{kingma2014adam}
D.~P. Kingma and J.~Ba, ``Adam: A method for stochastic optimization,''
  \emph{arXiv preprint arXiv:1412.6980}, 2014.

\bibitem{watanabe2018espnet}
\BIBentryALTinterwordspacing
S.~Watanabe, T.~Hori, S.~Karita, T.~Hayashi, J.~Nishitoba, Y.~Unno, N.~{Enrique
  Yalta Soplin}, J.~Heymann, M.~Wiesner, N.~Chen, A.~Renduchintala, and
  T.~Ochiai, ``{ESPnet}: End-to-end speech processing toolkit,'' in
  \emph{Proceedings of Interspeech}, 2018, pp. 2207--2211. [Online]. Available:
  \url{http://dx.doi.org/10.21437/Interspeech.2018-1456}
\BIBentrySTDinterwordspacing

\bibitem{hayashi2020espnet}
T.~Hayashi, R.~Yamamoto, K.~Inoue, T.~Yoshimura, S.~Watanabe, T.~Toda,
  K.~Takeda, Y.~Zhang, and X.~Tan, ``{Espnet-TTS}: Unified, reproducible, and
  integratable open source end-to-end text-to-speech toolkit,'' in
  \emph{Proceedings of IEEE International Conference on Acoustics, Speech and
  Signal Processing (ICASSP)}.\hskip 1em plus 0.5em minus 0.4em\relax IEEE,
  2020, pp. 7654--7658.

\end{thebibliography}


\end{document}